\documentclass[runningheads,a4paper]{llncs}
\usepackage{makeidx}  % allows for indexgeneration
\usepackage[misc,geometry]{ifsym}  %% aggiunto per mettere icona lettere vicino ad autore
\usepackage{graphicx} %% per le figure
\begin{document}
\frontmatter          % for the preliminaries
\pagestyle{headings}  % switches on printing of running heads
\addtocmark{Hamiltonian Mechanics} % additional mark in the TOC

\mainmatter              % start of the contributions
%
%\title{SynCE-NET: Synthetic Contrast Enhancement Network for  
%heart chambers estimation from non-contrast CT images}
\title{Synthetic contrast enhancement in cardiac CT with Deep Learning}
\titlerunning{Synthetic contrast enhancement in cardiac CT with Deep Learning}  % abbreviated title (for running head)
\author{Gianmarco Santini\inst{1,}\inst{2} \Letter , Lorena M. Zumbo\inst{1}, Nicola Martini\inst{2}, Gabriele Valvano\inst{3}, Andrea Leo\inst{3}, Andrea Ripoli\inst{2}, Francesco Avogliero\inst{2}, Dante Chiappino\inst{2}, Daniele Della Latta\inst{2}\Letter}
%\author{**** *****\inst{1}, ******* *******\inst{1,}\inst{2} \Letter , ****** *******\inst{2}, ****** *******\inst{3}, ****** ***\inst{3}, ****** %*****\inst{2}, ********* *********\inst{2}, ***** *********\inst{2}, ******* ********* \inst{2}}
%
\authorrunning{G. Santini et al.} % abbreviated author list (for running head)
%\authorrunning{***** et al.}
%Craig Chambers, Kim B. Bruce, and Elisa Bertino}
%
\institute{Department of Information Engineering, University of Pisa, Pisa, Italy\\
%\email{gianmarco.santini@ing.unipi.it}
\and Imaging department, Fondazione Gabriele Monasterio, Massa, Italy\\
\and
IMT Lucca\\
\email{gianmarco.santini@ing.unipi.it}\\
\email{dellalatta@ftgm.it}}
%\institute{************************************************\\
%\email{********************************}
%\and
%***********************************************\\
%\and
%**********************************************}

\maketitle              % typeset the title of the contribution

\begin{abstract}
In Europe, the 20\% of the CT scans cover the thoracic region. The acquired images contain information about the cardiovascular system that often remains latent due to  the lack of contrast in the cardiac area. On the other hand, the contrast enhanced computed tomography (CECT) represents an imaging technique that allows to easily assess the cardiac chambers volumes and the contrast dynamics.
With this work we aim to face the problem of extraction and presentation of these latent information, using a deep learning approach with convolutional neural networks.
Starting from the extraction of relevant features from the image without contrast medium, we try to re-map them on features typical of CECT, to synthesize an image characterized by an attenuation in the  cardiac chambers as if a virtually iodine contrast medium was injected. The purposes are to guarantee an estimation of the left cardiac chambers volume and to perform an evaluation of the contrast dynamics.
Our approach is based on a deconvolutional network trained on a set of 120 patients who underwent both CT acquisitions in the same contrastographic arterial phase and the same cardiac phase. To ensure a reliable predicted CECT image, in terms of values and morphology, a custom loss function is defined by combining two terms. The first contribute is an error function to find a pixel-wise correspondence, which takes into account the similarity in term of Hounsfield units between the input and output images. A second term is added to enforce the definition of the chambers. It's expressed by a cross-entropy computed on the binarized versions of the synthesized and of the real CECT image.
The proposed method is finally tested on 20 subjects; the left heart chambers are evaluated with the Dice metric ($0.88\pm0.03$) and the volume percentage error ($9.1\pm6.2$\%), while the dynamics of the x-ray attenuation is evaluated with the NMI index ($0.93\pm0.03$) and PSNR($48.30\pm2.03$).

%\keywords{Deep Learning, CT,  Convolutional Neural Network}
\end{abstract}
\section{Introduction}
In Europe the 20\% of the CT scans cover the thoracic region \cite{Radreport180}.
Improvements in CT scanner technology-specifically provide during routine chest examination, heart images much less degraded by cardiac motion artifacts and that allow detailed evaluation of the cardiac structures \cite{bruzzi2006and}. Therefore, the acquired images actually contain good quality information about the cardiovascular system that often remains latent, due to the lack of contrast in the cardiac area.\\
On the other hand, the contrast enhanced computed tomography (CECT) represents an imaging technique that allows to easily assess the cardiac chambers volumes and the contrast dynamics \cite{rizvi2015analysis}. In fact, from the clinical point of view, it is important to define the morphology of the cardiac chambers to identify any patients affected by cardiopathies or valvular pathologies.\\
A physician, viewing several CECT cardiac images, develops visual memories related to the contrast medium distribution in the cardiac chambers, updating and enriching those memories based on experience and on prior knowledge. As a matter of fact, the acquired expertise allows clinicians to transfer information about the shapes and positions of left atrium (LA) and the left ventricle (LV) onto an image where they are not visible, thanks to imagery and memory retrieval operations \cite{van2012schema}. \\
In computer science the deep convolutional neural networks (DCNN) architecture, inspired by the biology of the human visual system, has achieved breakthrough performance in image analysis. This has brought us to develop a DCNN model able to create a contrast enhanced image from a non-contrast enhanced  one.\\
In this work we demonstrate how employing a DCNN model it is possible to synthesized an image characterized by an attenuation in the left cardiac chambers as if a virtually iodine contrast medium was injected. In addition, by exploiting the DCNN capability to extract features with spatial and contrast invariance \cite{lecun1995convolutional}, we suggest that the designed model is able to mimic the human visual memory system outperforming an expert radiologist in the volumetric assessment of left heart chambers.
\section{Materials and Methods}
\subsection{Dataset}
The study includes 150 ECG triggered CT scans acquired during a standard cardiac CT session, with a tube voltage of 120 KVp and a modulated tube current (50-350mA). The images are reconstructed with a dimension of $512\times512$ pixels, a resolution ranging between 0.3 mm to 0.5 mm, and a fixed slice thickness of 3.0 mm. The CECT scans are obtained with tube voltage peak variable from 80-120 KVp, a modulated tube current(50-500mA) and successively reconstructed with the same dimensions and in plane resolution of the corresponding basal CT.\\
Both the acquisitions are acquired at the same cardiac telediastolic phase, i.e. 75\% R-R. All images are reconstructed using iterative filter for ionizing dose reduction.  For each patient 50 ml of contrast medium has been injected venously, at a flow rate of 4.5 ml/s. All volume are acquired in arterial contrastographic phase,  characterized by an higher quantity of contrast medium in the left part of the heart compared to the right one, allowing a better visualization of the left cardiac chambers and coronary vessels. For each CECT scans a manual segmentation of the left atrium (LA) and left ventricle (LV) is also provided by expert radiologists. This information is used, together with the CECT images, as ground truth for the network. 
\subsection{Preprocessing}
Before feeding the data into the model, some preprocessing operations are required.\\
To cope with some misalignments in the images, caused by respiratory motion, a rigid registration on CECT scans is performed as first processing step, using the CT images as references and the mutual information as cost function to minimize. \\
While the two acquisitions share the same isotropic in plane resolution, they have different slice thickness. For such reason we also perform a reslicing operation on the registered CECT scans, bringing the two acquisitions to a common axial resolution of 3.0 mm.\\
Finally, as our aim is to produce a synthetic contrastographic map capable to characterize the heart left chambers, instead of using the entire collected CT volume, we consider for each subject the only axial slices where the LA and the LV are present and have been manually segmented on the CECT cases; the remaining slices are instead discarded.
\begin{figure*}[htbp!]
      \centering
          \includegraphics[width=0.85\textwidth]{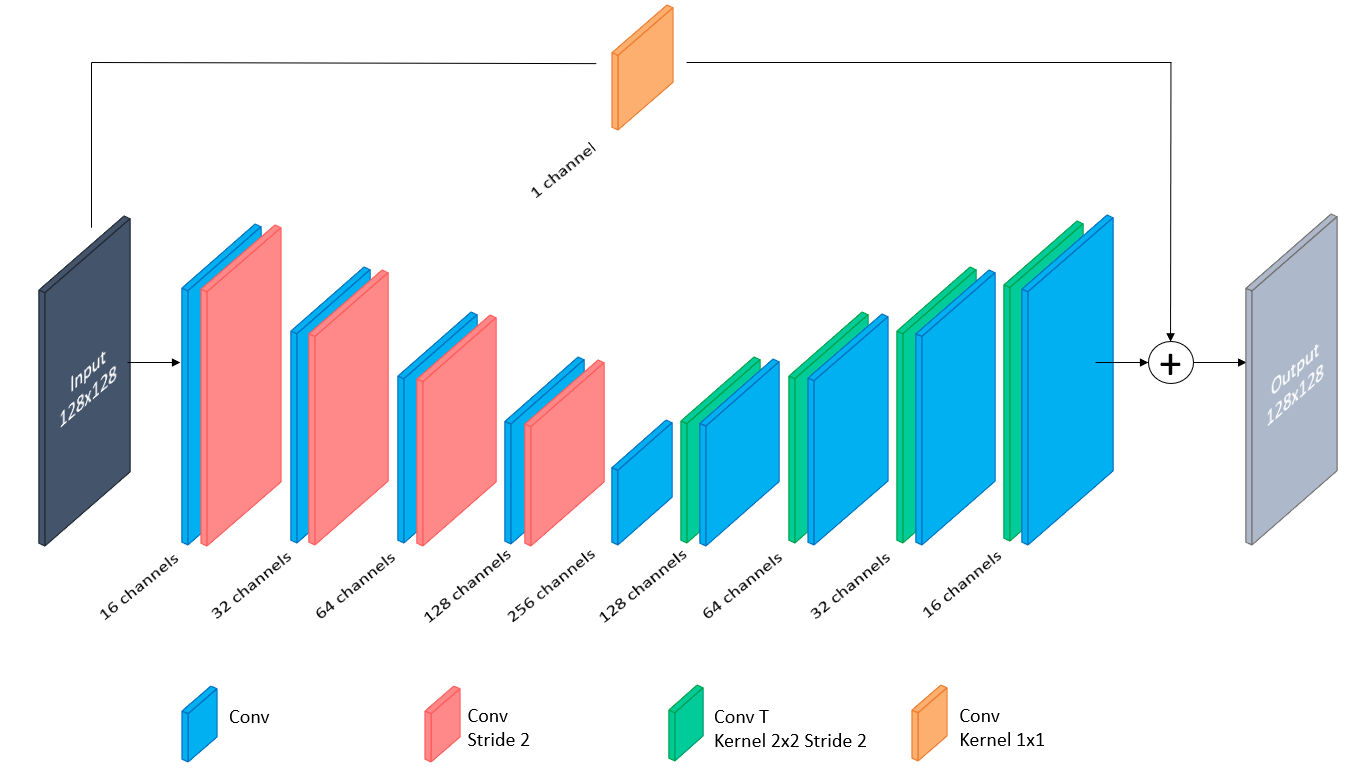}
      \caption{Block diagram of the network used.}
      \label{figure1}
\end{figure*}
\subsection{Learning phase}
\subsubsection{Model architecture:}
The architecture model adopted is a fully convolutional network, inspired by the well-known Deconvolutional Network \cite{noh2015learning}. Figure \ref{figure1} illustrates in detail the complete architecture designed, where a first encoding path is used as features extractor from the input image, while the decoding path progressively reconstructs the contrast enhanced image using the previous derived features. \\
The encoder comprises eight convolutional layers with a squared kernel size, equal to 3. For each of these layers a ReLU activation function is used, immediately after a batch normalization \cite{ioffe2015batch} applied to the convolution operation output. No max-pooling is used, in fact the downsampling operation is performed using a stride of 2 pixels in all the convolutional layers with even index. An intermediate convolutional layer separates the encoder from the decoder. Here the expanding module is symmetrically built using eight convolutional layers, where an up-sampling of factor 2 is realized this time on the odd layers, employing transposed convolutions with a $2\times2$ kernel size, stride 2 and no zero-padding.\\
Finally, a skip connection, with a $1\times1$ convolutional block is included in the model. Its output is added to the last convolutional layer where a single 3x3 kernel and a linear activation function are used, to generate continuous values for each pixels, instead of probability scores.  Such implementation is adopted with the precise aim to obtain a residual representation of the input \cite{he2016deep}, propagating spatial information from the input image that can get lost in the encoding process.
\subsubsection{Loss function:}
To correctly map the input CT image to the respective CECT image, and to recreate an adequate attenuation dynamics for the extraction of the left heart chambers, we propose a loss function given by the combination of two terms. 

\begin{equation}
  L(\tilde{y}, y) =  \left[\alpha \cdot RMSE(\tilde{y}, y) + \beta \cdot BCE(sig(\tilde{y}), y^{01})\right] \cdot H_{mask} + \frac{\lambda}{2} ||W||^2_2
 \label{eq1}
\end{equation}

\begin{equation}
  RMSE =  \sqrt{\frac{1}{N} \sum\limits_{i=0}^N (\tilde{y_i} - y_i)^2}  
   \label{eq2}   
 \end{equation}

\begin{equation}
  BCE =  -\sum\limits_{i=0}^N \left[y_i^{01} \cdot \log{(sig(\tilde{y_i})) + (1-y_i^{01})\cdot \log(1-sig(\tilde{y_i}))}\right]
   \label{eq3}
\end{equation}
\\
The first term is used to learn the network parameters by minimizing the difference between the synthesized image and the real CECT image. Specifically, it is a regressive term implemented with the root mean square error (RMSE) as in Eq. \ref{eq2} where $\tilde{y_i}$ and $y_i$ represent the predicted CECT image and the real CECT scan respectively, while $i$ iterates over the pixels.\\
The second term is the binary cross entropy (BCE) computed between the manually segmented chambers $y^{01}$ and the normalized version of the generated CECT image (see Eq. \ref{eq3}). The predicted output is in fact transformed in a binary image using a sigmoid function, modified to work as an approximation of a Heaviside step function: $sig(x) = \frac{1}{1+e^{-s(x - v_{th})}}$ . The variable $s$ controls the function steepness and is appropriately chosen to avoid exploding gradient effects. The $v_{th}$ is instead used to shift the sigmoid function and center it  around an appropriate value to discriminate the LA and the LV from the chamber walls and other structures. This threshold is taken considering the mean on the minimum HU values in the ground truth masks of the heart left chambers. The final cost function (Eq. \ref{eq1}) is the result of linear combination of the two terms and it is multiplied by a binary mask of the heart, extracted to focus the analysis only on the cardiac region, as consequence of the observation that the contrast features are mainly located in this region.
$\alpha$, $\beta$  as $v_{th}$  and $s$ are all considered model hyperparameters. An L2 regularization with a weight decay $\lambda$ is also added to prevent overfitting.
\subsection{Evaluation metrics}
\begin{figure*}[htbp!]
      \centering
          \includegraphics[width=0.90\textwidth]{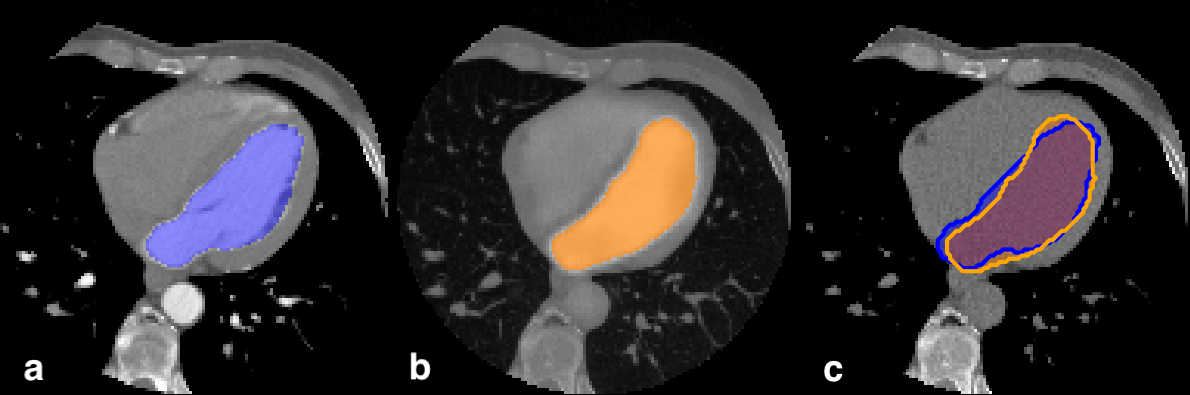}
      \caption{Left chambers quality evaluation. a) Ground truth mask on CECT image, b) estimation of the left chambers from synthesized CECT  c) overlapping region quantified by Dice metric.}
      \label{figure2} 
\end{figure*}

The output image quality is assessed with the Normalized Mutual Information index (NMI) and Peak Signal to Noise Ratio index (PSNR) as they allow to quantify the capability of the model to recreate a synthesized cardiac image as close as possible to the real CECT one.\\
The chambers estimation is evaluated with the Dice index. On each axial slice the overlapping grade between the heart chambers, segmented by threshold on the synthesized CECT image, and the manual segmentation is quantified. Additionally, the agreement of the predicted measurements with the manual references is highlighted by the Pearson coefficient ($\rho$) and the Bland-Altman plot, together with the volume percentage error ($dV\%$).\\
A set of qualitative comparisons are also performed to estimate the efficiency of the new measurement method on 10 images randomly chosen from the total scans in test set. Using the chambers ground truth as reference we compare the network results with those obtained by two expert radiologists, who have been asked to draw the cardiac chambers on the image without contrast. Dice index (see Fig.\ref{figure2}) and $dV\%$ are used to evaluate the aforementioned comparisons. A Pearson coefficient is instead computed to quantify the intra and the inter observer variability on two operators chosen for the above task. Clearly, they are not among those who provided the ground truth. \\
All the metrics and the comparisons are evaluated inside the region of interest, delimited by the heart mask extracted for each subject.
\section{Experimental settings and Results}
In this work, we employ a total number of patients equal to 150, that we randomly split in 120 cases to train the model, 10 cases for its validation and 20 cases for the performance evaluation. For each patient we have collected a CT and a CECT scan, the segmentation of the LA and the LV, and a binary mask of the heart. All of these images are resampled from the original dimensions to 128x128 pixels resulting in a loss of spatial information but a considerable acceleration in processing time.\\
We implemented the network in Tensorflow \cite{abadi2016tensorflow} and trained the model from scratch, launching it on a single NVIDIA TITAN X GPU machine for 800 epochs. We used the Adam optimization for training the network parameters, with a learning rate fixed to $10^{-4}$ and a batch size of 32 samples of 2D images from the CT volume axial projection.
To increase the training cases and guarantee a higher generalization power in the prediction phase, we also use data augmentation, applying random rotation with a max angle of 25 degrees on all the couple of input and output provided to the network. 
\begin{figure*}[htbp!]
      \centering
          \includegraphics[width=0.85\textwidth]{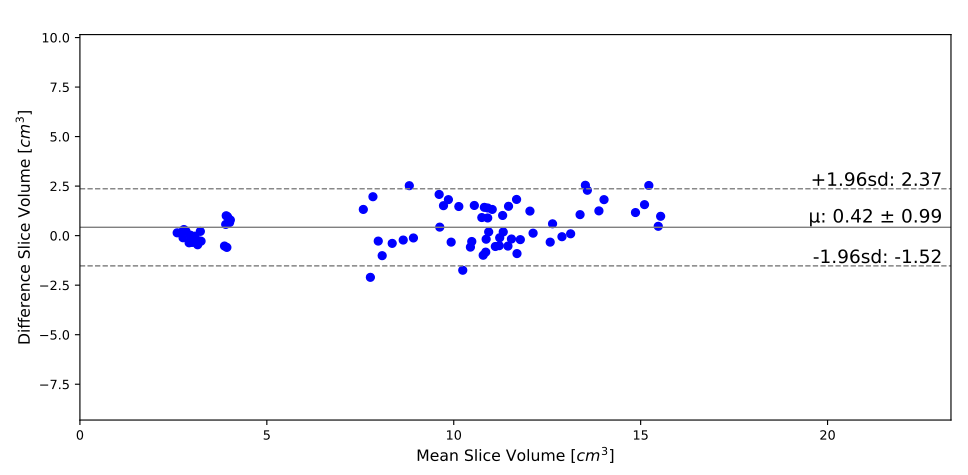}
      \caption{Bland-Altman plot.}
      \label{figure3}
\end{figure*}
About the loss hyper-parameters we set $\alpha$, $\beta$ and $\lambda$  to 1, 0.01 and 0.001 respectively, while for $s$ we have chosen the value of 10. Finally, we have found in 300 HU an adequate inferior limit for $v_{th}$ to discriminate LA and LV in the synthesized CECT image.\\
To evaluate the model performance 78 test slices have been extracted from 20 test volumes. With the above settings we achieve an average NMI value of $0.93 \pm 0.03$ and a PSNR of $48.30 \pm 2.03$ in the overall process of synthesis. Compared to the manually segmented LA and LV, our prediction reaches a mean Dice of $0.88\pm0.03$, while an high agreement with the chambers volumes is reported by a $\rho=0.97$ (p\textless 0.01), Bland-Altman plot (Fig. \ref{figure3}) and a $dV\%$ of $9.1 \pm 6.2$\%.\\
The qualitative comparison outcomes show how the model manages to create a volume by committing an error of 7\% ($dV\%$), which is close to the one committed by humans, but with the advantage of producing a more reliable geometry: $Dice_{Deep}=0.89\pm0.03$ vs $Dice_{Human}=0.85\pm0.05$. A good intra observer reproducibility ($\rho=0.99$) is also observed, while a lower inter observer reproducibility is instead reported ($\rho=0.85$).\\
\section{Discussion and Conclusion}
We propose a novel approach to synthesize cardiac CECT images from contrast-free CT thoracic scans, exploiting the ability of the DCNN to mimic the human visual system and to regenerate the imagery and memory retrieval operations.
\begin{figure*}[htbp!]
      \centering
          \includegraphics[width=0.99\textwidth, height=5.5cm]{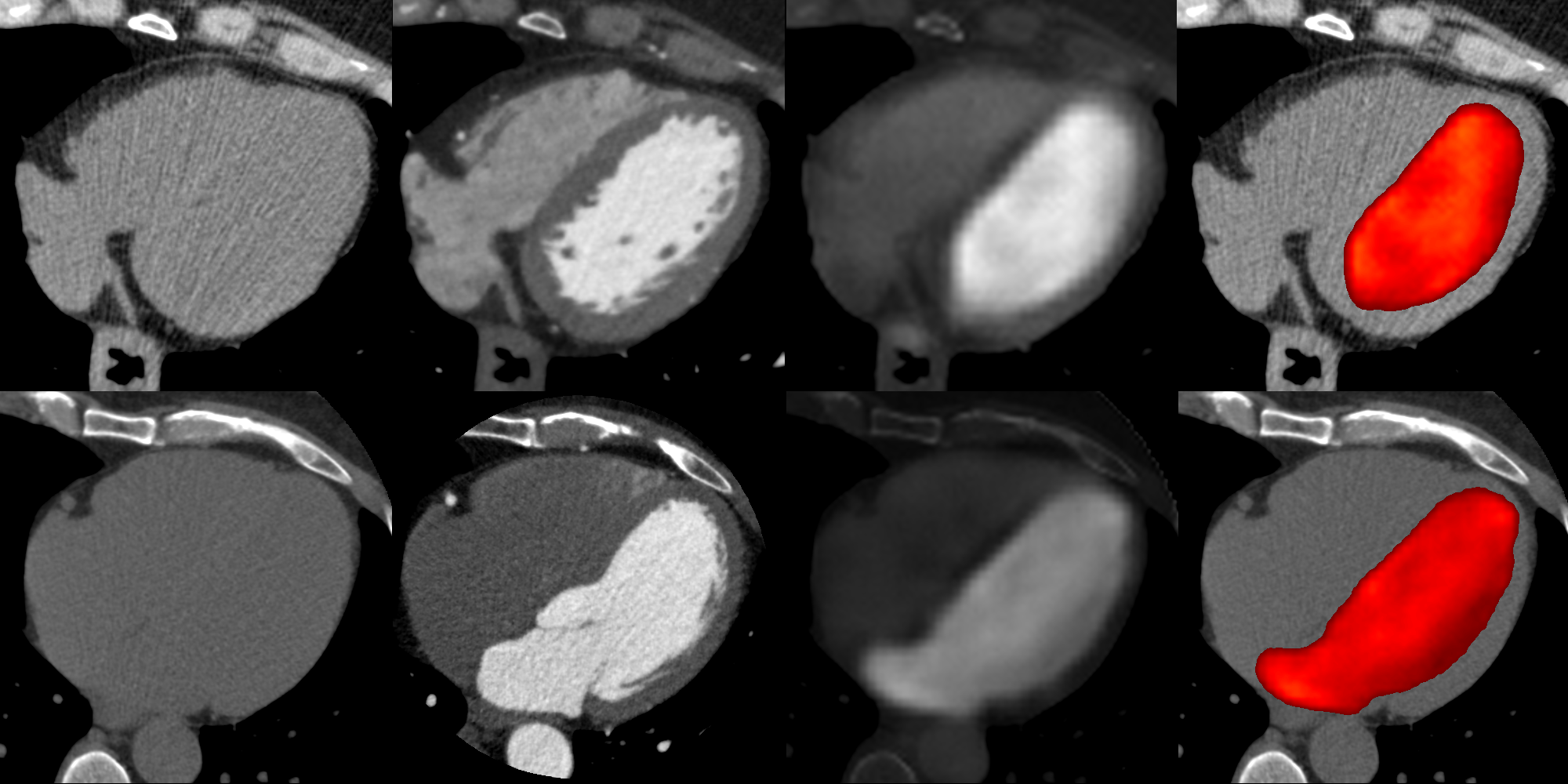}
      \caption{Two examples of image synthesizing. From left to right: the contrast free CT image, the real CECT image, the synthesized CECT image and the fusion of the basal CT image with the estimated chambers.}
      \label{figure4}
\end{figure*}
The synthetic CECT images show a good similarity and contrast dynamics ($NMI=0.93\pm0.03$, $PSNR=48.30\pm2.03$) compared to the gold standard (i.e a real CECT images),  allowing a simple extraction of the left cardiac chambers by thresholding.
The comparison of the automatic segmentation with the manual reference annotations highlights an independence of the synthesized chambers morphology from both the heart shapes and the slice positions in the CT volume (Dice=0.88), Fig.\ref{figure4}.\\
From the qualitative analysis it's then possible to assess how the DCNN is able to overcome the inter observer variability outperforming the human performances (intra observer correlation IAOC $= 0.99$, inter observer correlation IROC $= 0.85$) and offering the chance to have a fast and repeatable measurement.\\
From the presented results we can in fact assess the network capability to mimic the neurophysiological process of the image synthesis after the training phase. The DCNN properties make the model able to extract deep features and latent information which are not directly perceptible by the human eye, resulting in more accurate ($Dice_{Deep} \textgreater Dice_{Human}$) and reproducible measurements (IROC=$0.85$). Moreover, as consequence of a $dV\% = 9,1 \%$ on the entire test set, we can hypothesize to employ this method in clinical applications such as the identification of patients with heart or valvular diseases.\\
To conclude, we have proposed a DCNN approach to synthesize a contrast enhanced image from a basal CT scan, where it is possible to segment the left atrium and ventricle by threshold. What emerges is a promising approach that can offer  the chance to retrieve from patients, who undergo chest CT exams, volumetric information, otherwise hidden, about the cardiac chambers.

%\begin{thebibliography}{5}
%
%\bibitem {clar:eke}
%Clarke, F., Ekeland, I.:
%Nonlinear oscillations and
%boundary-value problems for Hamiltonian systems.
%Arch. Rat. Mech. Anal. 78, 315--333 (1982)

%\bibitem {clar:eke:2}
%Clarke, F., Ekeland, I.:
%Solutions p\'{e}riodiques, du
%p\'{e}riode donn\'{e}e, des \'{e}quations hamiltoniennes.
%Note CRAS Paris 287, 1013--1015 (1978)

%\bibitem {mich:tar}
%Michalek, R., Tarantello, G.:
%Subharmonic solutions with prescribed minimal
%period for nonautonomous Hamiltonian systems.
%J. Diff. Eq. 72, 28--55 (1988)

%\bibitem {tar}
%Tarantello, G.:
%Subharmonic solutions for Hamiltonian
%systems via a $\bbbz_{p}$ pseudoindex theory.
%Annali di Matematica Pura (to appear)

%\bibitem {rab}
%Rabinowitz, P.:
%On subharmonic solutions of a Hamiltonian system.
%Comm. Pure Appl. Math. 33, 609--633 (1980)

%\end{thebibliography}


\bibliographystyle{splncs03}
\bibliography{biblio}

\begin{thebibliography}{1}
\providecommand{\url}[1]{\texttt{#1}}
\providecommand{\urlprefix}{URL }

\bibitem{Radreport180}
https://ec.europa.eu/energy/sites/ener/files/documents/{RP}180.pdf. Tech. rep.
  (2014)

\bibitem{abadi2016tensorflow}
Abadi, M., Barham, P., Chen, J., Chen, Z., Davis, A., Dean, J., Devin, M.,
  Ghemawat, S., Irving, G., Isard, M., et~al.: Tensorflow: A system for
  large-scale machine learning. In: OSDI. vol.~16, pp. 265--283 (2016)

\bibitem{bruzzi2006and}
Bruzzi, J.F., R{\'e}my-Jardin, M., Delhaye, D., Teisseire, A., Khalil, C.,
  R{\'e}my, J.: When, why, and how to examine the heart during thoracic ct:
  Part 1, basic principles. American Journal of Roentgenology  186(2),
  324--332 (2006)

\bibitem{he2016deep}
He, K., Zhang, X., Ren, S., Sun, J.: Deep residual learning for image
  recognition. In: Proceedings of the IEEE conference on computer vision and
  pattern recognition. pp. 770--778 (2016)

\bibitem{ioffe2015batch}
Ioffe, S., Szegedy, C.: Batch normalization: Accelerating deep network training
  by reducing internal covariate shift. In: International conference on machine
  learning. pp. 448--456 (2015)

\bibitem{van2012schema}
van Kesteren, M.T., Ruiter, D.J., Fern{\'a}ndez, G., Henson, R.N.: How schema
  and novelty augment memory formation. Trends in neurosciences  35(4),
  211--219 (2012)

\bibitem{lecun1995convolutional}
LeCun, Y., Bengio, Y., et~al.: Convolutional networks for images, speech, and
  time series. The handbook of brain theory and neural networks  3361(10),
  1995 (1995)

\bibitem{noh2015learning}
Noh, H., Hong, S., Han, B.: Learning deconvolution network for semantic
  segmentation. In: Proceedings of the IEEE International Conference on
  Computer Vision. pp. 1520--1528 (2015)

\bibitem{rizvi2015analysis}
Rizvi, A., Dea{\~n}o, R.C., Bachman, D.P., Xiong, G., Min, J.K., Truong, Q.A.:
  Analysis of ventricular function by ct. Journal of cardiovascular computed
  tomography  9(1),  1--12 (2015)

\end{thebibliography}

%\clearpage
%\addtocmark[2]{Author Index} % additional numbered TOC entry
%\renewcommand{\indexname}{Author Index}
%\printindex
%\clearpage
%\addtocmark[2]{Subject Index} % additional numbered TOC entry
%\markboth{Subject Index}{Subject Index}
%\renewcommand{\indexname}{Subject Index}
%\input{subjidx.tex}
\end{document}